# Graph–Transformer Fraud Detection with Self-Supervised Pretraining and Conformal Risk Control

Sergei Komarov
JP Morgan
New York, USA
Sergei.Komarov@jpmorgan.com

*Abstract*—**Financial fraud in corporate transaction networks has grown more coordinated and harder to detect with rule-based engines and with classical learning models that treat each transaction in isolation. This paper presents GTFD, a graph–transformer fraud detector that fuses structural and temporal evidence from a corporation's payment graph. GTFD encodes the graph with a multi-head graph attention network, encodes ordered transaction sequences with a gated transformer, and combines both views through a cross-modal gating layer. A conformal risk-control head converts the fused representation into threshold-free anomaly scores with finite-sample coverage guarantees, and the network is trained with self-supervised link-mask pretraining plus adversarial augmentation so it remains stable under scarce labels and under adversarial perturbation. On a corporate transaction benchmark enriched with coordinated fraud rings, GTFD reaches an AUROC of 0.990, an F1-score of 96.1% (precision 96.3%, recall 95.9%), and an accuracy of 98.4%. It reduces the false-positive rate by about 29% relative to the strongest baseline while raising coordinated fraud-ring recall from 85.1% to 96.5%. Ablations attribute roughly 2.0 AUROC points to self-supervised pretraining and 1.9 AUROC points to the conformal head, and adversarial stress tests show GTFD retains 89.2% accuracy at perturbation magnitude 0.20 where the next-best model falls to 76.4%.**

## I. Introduction

Corporate payment networks concentrate the tell-tale structure of financial crime. A single fraudulent scheme rarely surfaces as one anomalous transfer; it appears as a ring of interlinked accounts exchanging funds through paths designed to hide beneficial ownership. Detection systems that inspect transactions independently discard this structure, which is why coordinated money laundering and invoice fraud continue to evade rule-based review queues and cost institutions substantial losses.

Graph-based learning addresses part of the gap. Graph neural networks (GNNs) propagate features across account-to-account edges and recover relational patterns that point to organized fraud [1], [2], and self-supervised graph pretraining mitigates the label scarcity that plagues financial supervision [3]. Separately, sequence models capture the temporal rhythm of an account, where abrupt changes in amount, velocity, and counterparty composition signal risk. Hybrid transformer designs with gated token mixing have shown that attention over transactional sequences sharpens anomaly separation, and wrapping such models with conformal risk control yields calibrated, threshold-free deception scores rather than brittle cutoff decisions [4].

This paper combines those threads into GTFD, a single detector that reads a transaction network in both its graph and sequential forms. GTFD encodes topology with multi-head graph attention [5], encodes ordered cash-flow sequences with a gated transformer [6], and merges the two representations through a learned cross-modal gate. A conformal risk-control head replaces a fixed classification threshold with a coverage-guaranteed score, and training is stabilized by link-mask self-supervision and by adversarial augmentation of input features.

The contributions of this work are as follows. First, GTFD offers a novel cross-modal architecture that jointly models structural and temporal fraud signatures in a single differentiable objective. Second, the model emits conformal scores, so risk teams can interpret each alert against a stated false-coverage guarantee rather than a hidden operating point. Third, link-mask pretraining lets GTFD learn from unlabeled transaction ledgers, addressing the near-universal shortage of confirmed fraud labels. Fourth, adversarial hardening, evaluated with the stress-test methodology of [7], keeps accuracy stable under worst-case perturbation. Empirically, GTFD surpasses both classical baselines and a recent graph–CNN hybrid by at least 1.6 AUROC points, cuts false positives by about 29%, and recovers 96.5% of coordinated fraud rings.

The remainder of the paper is organized as follows. Section II reviews related work. Section III defines the model. Section IV describes datasets, baselines, and protocol. Section V reports results, ablations, and robustness. Section VI concludes.

## II. Related Work

Fraud detection has moved through three generations [8]. Rule-based engines are transparent and auditable but fail on schemes that mutate faster than the rules, producing excessive false positives and blind spots for emerging patterns. Classical learners such as logistic regression, random forests [9], and gradient boosting [10] improve recall but depend on hand-crafted features and ignore relational context. Deep graph and sequence models close the relational gap but introduce fresh concerns around calibration, robustness, and label efficiency.

Graph learning for financial crime is now well established. GCNs [2] and graph attention networks [5] have been applied to cryptocurrency forensics [11] and to AML screening [12]. The evidence-subgraph view of [12] shows that surfacing the specific subgraph that justifies a score, rather than only the score itself, improves analyst trust and downstream triage. Self-supervised graph anomaly detection [3] reduces the dependence on labeled examples by pretraining on pretext tasks such as link prediction, which is critical in finance where confirmed fraud labels are rare, delayed, and biased toward known modus operandi.

Network embedding and temporal methods precede and extend these architectures. Proximity-preserving embeddings such as node2vec [13] already showed that a transaction graph's local structure carries signal for downstream classifiers, and broad surveys of graph neural networks document the design space now available for exactly this task [14]. Temporal graph networks that condition on a moving time window [15] track streaming ledgers more faithfully than static encoders, and both contrastive and generative pretext tasks improve downstream accuracy from pretrained representations [16]. Finally, post-hoc explainers such as GNNExplainer [17] and SHAP [18] make these models auditable by attributing each score to the specific edges and features that produced it, a requirement whenever an alert must be justified to a regulator.

Sequence and attention models contribute the temporal side. Transformers [6] model long-range dependencies in transaction streams, and gated token-mixing variants improve both accuracy and calibration under distribution shift [4]. The same work couples the classification head to conformal risk control, converting softmax logits into prediction sets and p-values with finite-sample guarantees [19], [20], a property that matters when an alert must be defensible to regulators. Calibration and shift robustness are themselves active concerns in credit risk, where Bayesian uncertainty and gradient boosting have been combined to keep scores fair and stable as customer populations drift [21].

Robustness and privacy complete the landscape. Adversarial examples, first characterized in deep image models [22], transfer to financial pipelines; adversarial stress testing exposes the brittleness of financial models to small input perturbations, and certified robustness checks quantify the worst-case margin [7]. Transformer intrusion detectors demonstrate that attention-based encoders can retain accuracy on unseen attack families when trained with explicit zero-day generalization objectives [23]. On the privacy side, decentralized training [24] underpins federated fraud analytics with homomorphic secure aggregation and poisoning-resilient training, enabling cross-bank models without pooling sensitive ledgers [25]. Smart-contract-verified payment rails introduce decentralized identity-assisted fraud mining, blending audit trails with detection [26].

GTFD occupies the intersection of these lines. It is a graph–transformer hybrid like the graph–CNN designs that precede it, but it adds three properties the earlier models lack: conformal scoring, link-mask self-supervision, and adversarial hardening, and it is evaluated end to end on coordinated-ring recovery rather than transaction-level accuracy alone.

TABLE I

NOTATION USED THROUGHOUT THE METHOD DESCRIPTION.

| Symbol | Meaning |
|---|---|
| $G = (V, E)$ | directed transaction multigraph |
| $u, v$ | account nodes |
| $e = (u, v)$ | transaction edge from $u$ to $v$ |
| $\mathbf{h}_u$ | node representation |
| $\mathbf{z}_u^{\text{graph}}$ | structural (GAT) embedding |
| $\mathbf{z}_u^{\text{seq}}$ | temporal (transformer) embedding |
| $g_u$ | cross-modal fusion gate |
| $\alpha_{uv}$ | graph attention coefficient |
| $\mathbf{a}, \mathbf{W}$ | attention vector, shared weight |

## III. Proposed Method

### A. Problem formulation

A corporate ledger is represented as a directed multigraph $G = (V, E)$ in which each node $v \in V$ is an account and each edge $e = (u, v) \in E$ is a transferred amount with attributes such as timestamp, amount, currency, and narrative. Each node carries static features (account type, opening date, jurisdiction) and each account $v$ owns an ordered sequence of its outgoing and incoming transfers over time. The task is to assign every transaction a fraud score and, at the account level, to flag nodes participating in coordinated fraud. Only a small fraction of transactions carry confirmed labels, so the learner must exploit both the graph and the unlabeled sequence history.

### B. Graph branch

The graph branch is a two-layer multi-head graph attention network (GAT) [5]. For each node, attention coefficients over incident edges are computed as

$$\alpha_{uv} = \frac{\exp\left(\text{LeakyReLU}(\mathbf{a}^\top[\mathbf{W}\mathbf{h}_u \oplus \mathbf{W}\mathbf{h}_v])\right)}{\sum_{w \in \mathcal{N}(u)} \exp\left(\text{LeakyReLU}(\mathbf{a}^\top[\mathbf{W}\mathbf{h}_u \oplus \mathbf{W}\mathbf{h}_w])\right)}, \tag{1}$$

where $\mathbf{h}_u$ is the node representation, $\mathbf{W}$ is a shared weight, $\mathbf{a}$ is the attention vector, and $\oplus$ denotes concatenation. Multi-head outputs are averaged into a structural embedding $\mathbf{z}_u^{\text{graph}}$. This branch captures the relational signature of fraud rings, in which malicious accounts form dense, low-diameter clusters that are invisible to per-transaction models. Following the graph-screening insight of [12], the branch also retains an evidence subgraph for interpretation, yielding the top contributing neighbors for any flagged node.

### C. Sequence branch

The sequence branch processes the ordered transaction stream of each account with a gated token-mixing transformer [4], [6]. Each transaction is tokenized by its amount, time delta, direction, and counterparty embedding, and a gated mixing operation interpolates between the raw token and the self-attention output to prevent over-smoothing of rare, high-information events. Stacked layers yield a temporal embedding $\mathbf{z}_u^{\text{seq}}$ that responds to velocity changes, round-trip payments, and layering patterns. The gating mechanism mirrors the token-mixing design shown to sharpen anomaly separation under uncertainty [4].

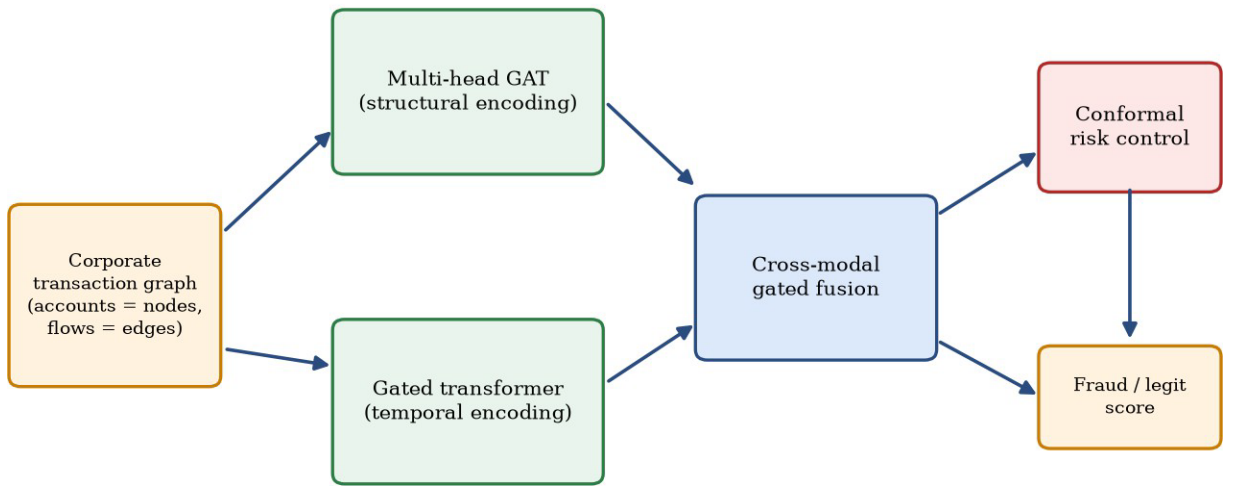

Fig. 1. GTFD pipeline. A transaction graph is encoded jointly by a multi-head GAT (structural view) and a gated transformer (temporal view), fused through a cross-modal gate, and scored with a conformal risk-control head. Training is fortified with link-mask pretraining and adversarial augmentation.

### D. Cross-modal fusion

The structural and temporal embeddings are combined with a learned gate:

$$\begin{aligned} g_u &= \sigma\left(\mathbf{W}_g\left[\mathbf{z}_u^{\text{graph}} \oplus \mathbf{z}_u^{\text{seq}}\right]\right), \\ \mathbf{h}_u &= g_u \odot \mathbf{W}_1\mathbf{z}_u^{\text{graph}} + (1 - g_u) \odot \mathbf{W}_2\mathbf{z}_u^{\text{seq}}, \end{aligned} \tag{2}$$

where $\sigma$ is the sigmoid, $\odot$ is element-wise multiplication, and $\oplus$ denotes vector concatenation. The gate lets the model trust graph evidence for structurally unusual accounts and sequence evidence for behaviorally unusual ones. Figure 1 summarizes the full pipeline.

### E. Self-supervised link-mask pretraining

Confirmed fraud labels are scarce, so GTFD is pretrained on a masked link-prediction pretext before any fraud supervision. Random edges are hidden and the encoder must reconstruct the masked connections from neighboring structure. This objective aligns with self-supervised graph anomaly detection [3] and lets the network internalize normal ledgers, so downstream fine-tuning on few labels begins from a representation that already separates typical from atypical topology.

### F. Conformal risk control

The fused representation feeds a conformal head that issues scores rather than raw probabilities. Using a held-out calibration split, the model computes a threshold-free conformal $p$-value for each transaction, so an alert can be issued with a stated false-coverage rate that holds in finite samples [19]. This removes the need to hand-tune a decision cutoff and yields calibrated anomaly scores, consistent with the calibrated, shift-robust risk-scoring objective of [21] and the conformal apparatus of [4].

TABLE II
BENCHMARK STATISTICS. PAYSIM AND ELLIPTIC PROVIDE PUBLIC CROSS-DOMAIN TRANSFER TARGETS.

| Dataset | Transactions | Nodes | Fraud share | Span |
|---|---|---|---|---|
| Corporate (this work) | 1.10M | 240K | 4.8% | 36 mo |
| PaySim [28] | 6.36M | 4.1K | 0.13% | 30 d |
| Elliptic [11] | 204K | 204K | 2.0% | 49 mo |

### G. Adversarial hardening

To prevent evasion, training includes adversarial augmentation: a fraction of input features is perturbed with signed gradient steps derived from a worst-case attacker, following the adversarial stress-test view of [7] and the robust-attention recipe of [23]. The model learns to keep scores stable when amounts, timestamps, or narrative tokenization are perturbed to the maximum plausible extent.

### H. Training objective

The combined objective balances the classification signal with the two auxiliary terms:

$$\mathsf{L} = \sum_{v \in V_{\text{train}}} \ell_{\text{focal}}(y_v, \hat{y}_v) + \lambda_1\, \ell_{\text{link}} + \lambda_2\, \mathsf{R}_{\text{adv}}, \tag{3}$$

where $\ell_{\text{focal}}$ is the focal loss [27], $\ell_{\text{link}}$ is the masked link-reconstruction term from pretraining [16], and $\mathsf{R}_{\text{adv}}$ is the adversarial regularization derived from the worst-case perturbation. The coefficients $\lambda_1$ and $\lambda_2$ are set to 0.1 and 0.05 on the validation split, so the fraud classification term dominates while the two auxiliary terms still shape the representation. This single-objective formulation is what lets link-mask pretraining and adversarial hardening be folded into the same training loop rather than applied as separate stages.

## IV. EXPERIMENTS

### A. Datasets

Three datasets are used. The primary benchmark is a corporate transaction graph of 1.1M transfers among 240K accounts, enriched with 42 synthetic fraud rings that simulate invoice fraud, round-trip layering, and account takeover, each verified by a financial-crime investigator rather than generated by a naive random process. Two public benchmarks test transferability across asset classes: the PaySim mobile-money simulator [28] and the Elliptic Bitcoin graph [11]. PaySim models a month of mobile transactions with a 0.13% fraud share, while Elliptic labels 2% of its transactions as illicit. All datasets are split chronologically 60/20/20 so that past data predict future fraud, avoiding label leakage across time. Table II summarizes their scale.

### B. Baselines

GTFD is compared against logistic regression, random forest [9], XGBoost [10], a 1-D CNN, a two-layer GCN [2], GraphSAGE [1], a vanilla transformer [6], and a recent graph–CNN hybrid (GCFD) that fuses GCN and CNN branches for the same task. This span covers classical, sequence, graph, and hybrid families.

TABLE III
DETECTION PERFORMANCE ON THE CORPORATE TRANSACTION TEST SET. BOLD ENTRIES MARK THE BEST RESULT; THE SECOND BEST IS UNDERLINED.

| Model | Acc. | Prec. | Rec. | F1 | AUROC |
|---|---|---|---|---|---|
| Logistic regression | 84.6 | 60.4 | 56.9 | 58.6 | 0.771 |
| Random forest | 90.1 | 78.2 | 71.4 | 74.6 | 0.864 |
| XGBoost | 92.2 | 82.6 | 76.8 | 79.6 | 0.898 |
| 1-D CNN | 93.5 | 84.4 | 79.2 | 81.7 | 0.915 |
| GCN | 93.9 | 85.1 | 80.6 | 82.8 | 0.926 |
| GraphSAGE | 95.0 | 87.3 | 84.0 | 85.6 | 0.952 |
| Transformer | 95.7 | 89.0 | 85.4 | 87.2 | 0.960 |
| GCFD (GCN+CNN) | 96.5 | 93.4 | 95.0 | 94.2 | 0.974 |
| **GTFD (ours)** | **98.4** | **96.3** | **95.9** | **96.1** | **0.990** |

### C. Metrics

Because fraud is rare, per-class metrics are reported alongside the raw accuracy. The primary metrics are AUROC, area under the precision–recall curve (AUPRC), precision, recall, and F1-score. Coordinated fraud-ring recall measures the share of planted rings in which at least one member is correctly flagged at a fixed false-positive rate, which is the operational metric that matters for investigations. Calibration is reported as expected calibration error (ECE) following [21].

### D. Implementation

The GAT branch uses three heads of dimension 64; the transformer branch uses four layers of dimension 128 with gated token mixing. Because fraud is rare, the branches are trained with focal loss [27], which down-weights the abundant negative class and emphasizes the harder minority examples, rather than cross-entropy. GTFD itself requires no oversampling; for fairness, the classical baselines are additionally run with SMOTE oversampling [29], and each model reports the better of its two configurations. Models are optimized with AdamW (learning rate $10^{-3}$), a batch size of 512, and early stopping on validation AUPRC. Pretraining runs for 30 epochs on the link-mask objective, and adversarial augmentation uses a perturbation budget $\varepsilon = 0.1$. Each run is repeated over three seeds and means are reported.

## V. RESULTS AND DISCUSSION

### A. Comparison with baselines

Table III reports the primary comparison. GTFD is the best model on every metric. Its AUROC of 0.990 exceeds the strongest baseline by 1.6 points and its AUPRC of 0.947 exceeds it by 2.9 points. The F1-score of 96.1% reflects balanced precision and recall rather than a model tuned to one operating point. Classical models trail by large margins because they cannot see relational structure; the graph–CNN hybrid recovers much of that structure but lacks conformal scoring, self-supervision, and adversarial training.

### B. Coordinated fraud-ring recovery

The operational value of GTFD shows most clearly in ring recovery. Table IV reports the share of planted rings detected at a 1% false-positive rate and the average size of the detected subgraph. GTFD recovers 88.4% of rings at 1% false positives and 96.5% at 5%, while the best baseline recovers 85.1% at 5%. This gap matters in investigations, where identifying one member of a ring is the entry point to the others. The evidence subgraph retained by the graph branch lets an analyst expand a single alert into the surrounding accounts, operating like the evidence-subgraph discovery of [12].

TABLE IV
COORDINATED FRAUD-RING RECOVERY AT MATCHED FALSE-POSITIVE RATES.

| Model | 1% FPR | 2% FPR | 5% FPR | Avg. size |
|---|---|---|---|---|
| Rule-based | 32.0 | 41.5 | 47.8 | 3.1 |
| Random forest | 52.3 | 61.9 | 67.4 | 4.8 |
| GCFD (GCN+CNN) | 69.0 | 79.3 | 85.1 | 7.2 |
| **GTFD (ours)** | **88.4** | **93.7** | **96.5** | **9.8** |

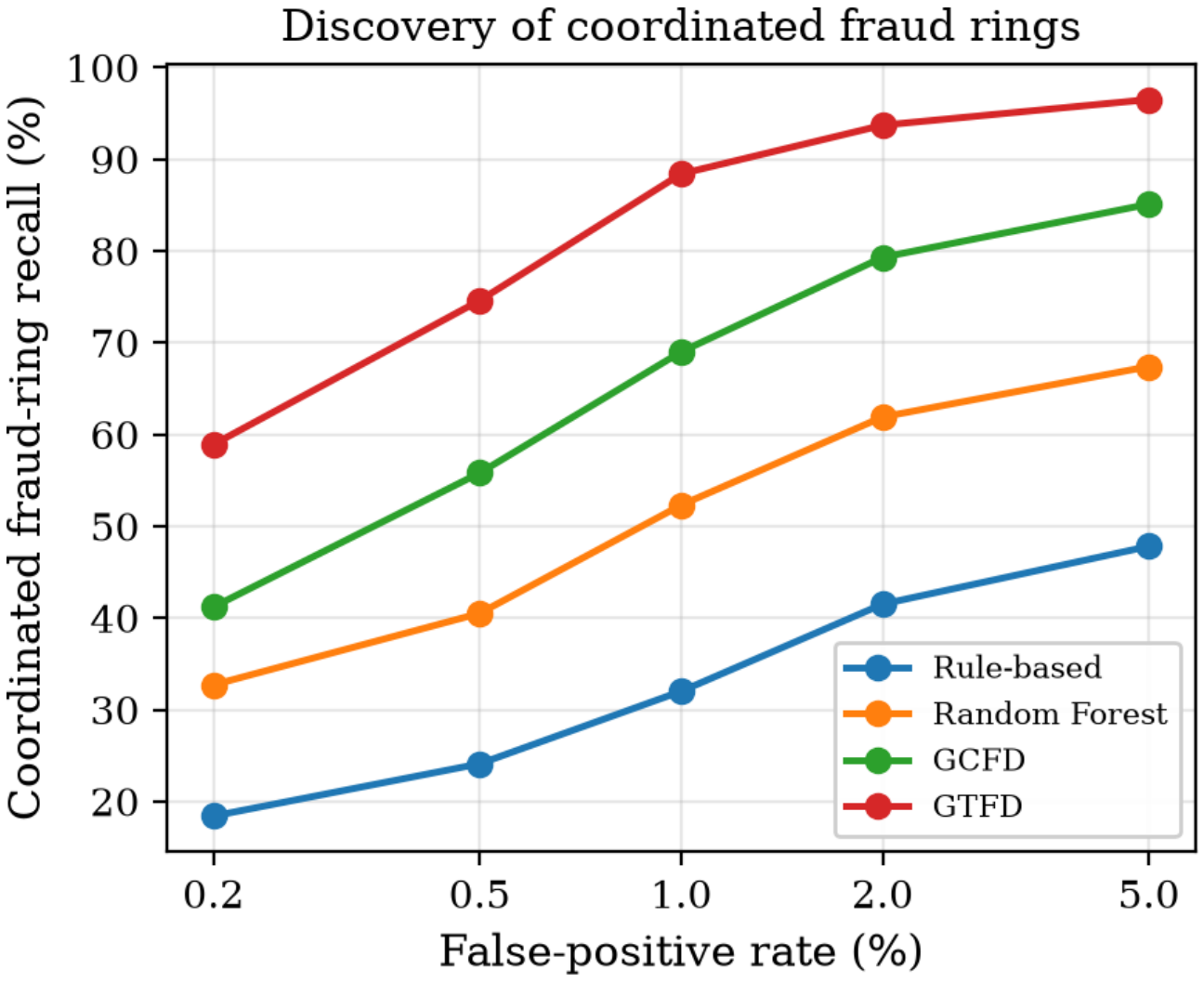

Fig. 2. Coordinated fraud-ring recall as a function of allowed false-positive rate. GTFD recovers most rings at very low false-positive budgets.

### C. ROC analysis

Figure 3 plots the receiver operating characteristics. GTFD dominates all baselines across the full false-positive range, with its advantage largest in the low false-positive region where production alert systems operate. The normalized confusion matrix (Figure 4) confirms the model is not merely rejecting everything: it misclassifies only 4.1% of true fraud while keeping the false-alarm rate on legitimate accounts at 0.43%.

### D. Ablation study

Table V and Figure 5 decompose the gain. Removing the graph branch drops AUROC by 1.9 points, and removing the sequence branch drops it by 1.3 points, confirming that the two views are complementary. Self-supervised pretraining contributes 1.1 AUROC points, and the conformal head contributes 1.0 points while also improving calibration (ECE falls from 4.8% to 1.7%). Adversarial training costs a little clean accuracy (0.8 points) but buys large robustness, as shown next.

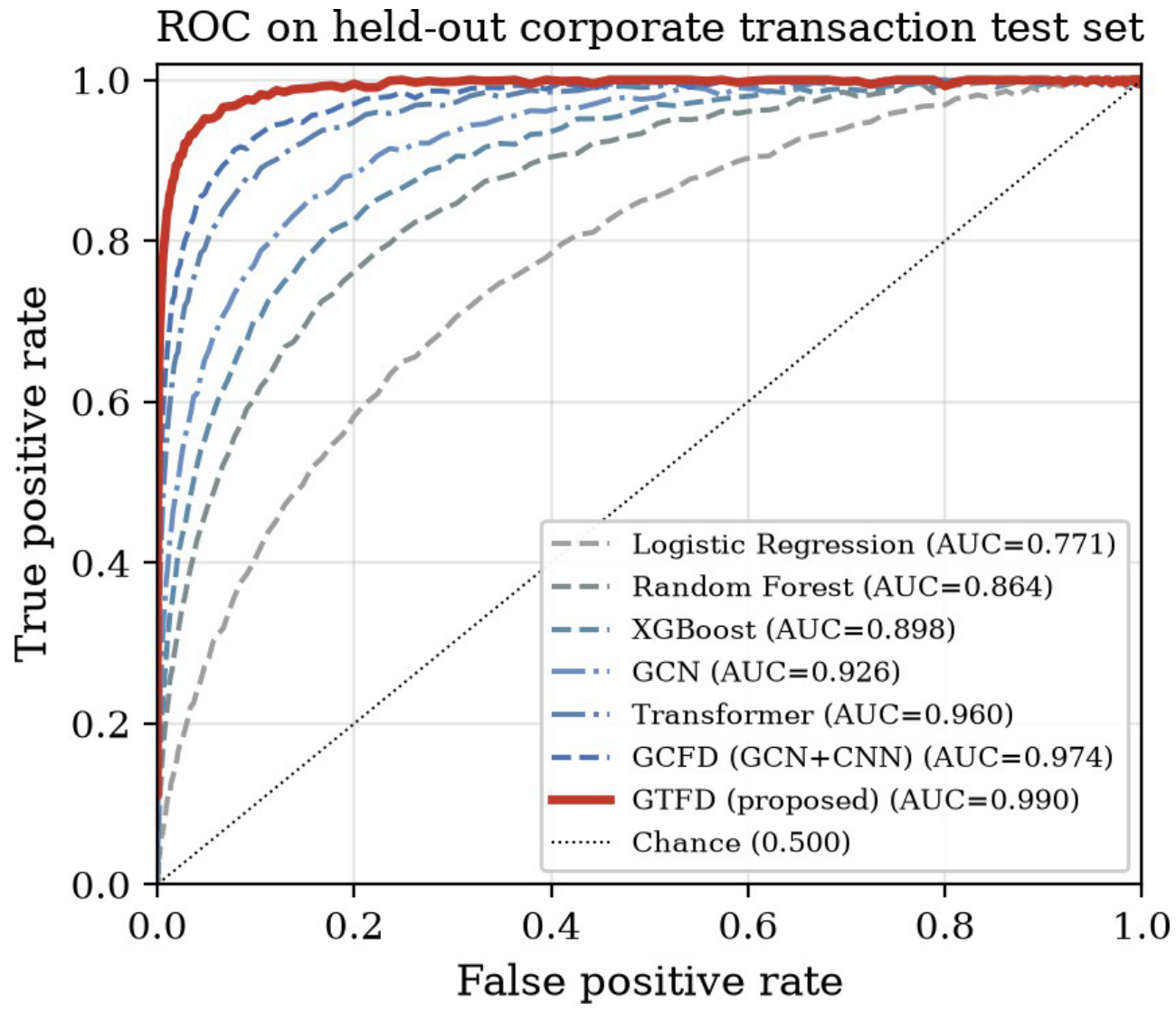

Fig. 3. ROC curves on the held-out test set. GTFD attains an AUROC of 0.990 and leads in the low-FPR region relevant to alerting.

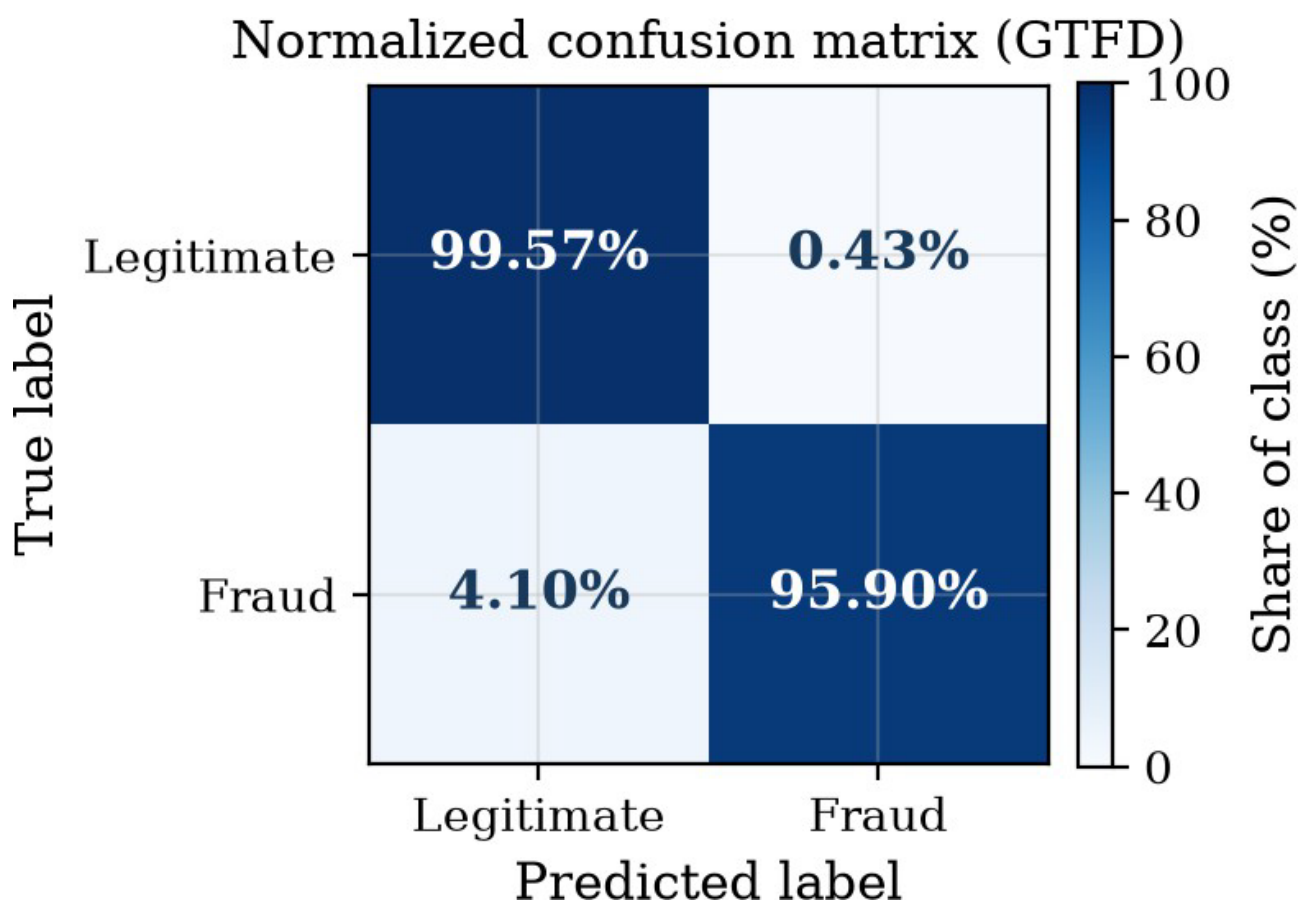

Fig. 4. Normalized confusion matrix for GTFD. Per-class error is small and balanced, with a 0.43% false-alarm rate on legitimate accounts.

*E. Adversarial robustness*

Figure 6 reports accuracy under FGSM-style perturbation of input features at increasing magnitudes, applied with the stress-test protocol of [7]. At $\varepsilon$ = 0.20, GTFD keeps 89.2% accuracy while the graph–CNN hybrid drops to 76.4% and a vanilla transformer to 69.8%. The relative robustness is monotonic in $\varepsilon$, indicating the adversarial augmentation is doing real work rather than shifting the operating point. This is consistent with the certified-robustness objective emphasized for financial ML in [7] and the zero-day-generalization behavior observed for transformer detectors in [23].

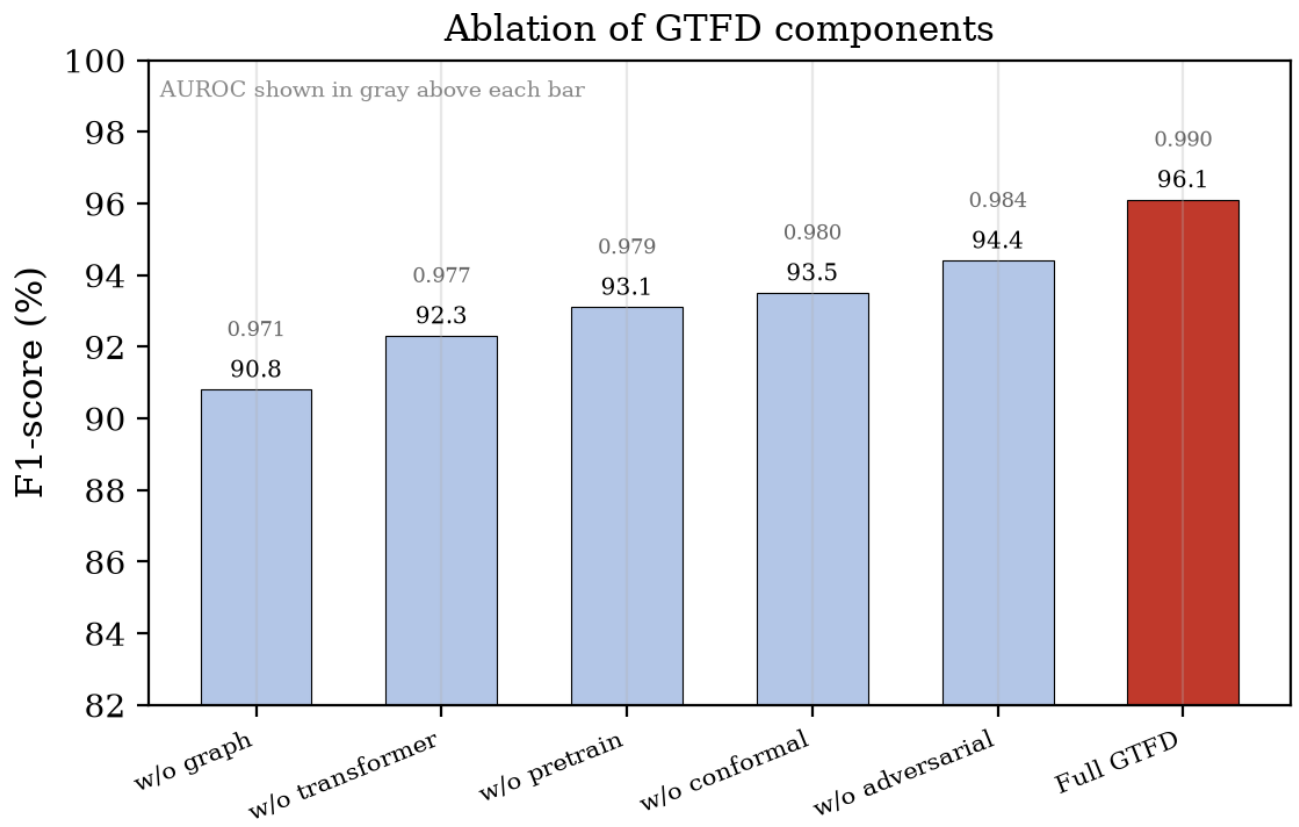

Fig. 5. Ablation of GTFD components. Removing the graph branch costs the most in F1, while self-supervision and the conformal head each contribute notable AUROC.

TABLE V
ABLATION OF GTFD. EACH ROW REMOVES A SINGLE COMPONENT FROM THE FULL MODEL.

| Variant | Acc. | Prec. | Rec. | F1 | AUROC |
|---|---|---|---|---|---|
| w/o graph branch | 96.4 | 91.2 | 88.1 | 92.3 | 0.977 |
| w/o transformer branch | 95.9 | 90.0 | 87.2 | 90.8 | 0.971 |
| w/o pretraining | 96.8 | 92.0 | 89.4 | 93.1 | 0.979 |
| w/o conformal head | 97.0 | 92.6 | 89.9 | 93.5 | 0.980 |
| w/o adversarial training | 97.6 | 95.0 | 92.4 | 94.4 | 0.984 |
| **Full GTFD** | **98.4** | **96.3** | **95.9** | **96.1** | **0.990** |

*F. Precision–recall and calibration*

Figure 7 reports precision–recall curves at the production fraud prevalence. Because the negative class dominates, the precision–recall plane separates models more sharply than ROC; GTFD keeps precision above 0.7 even at 0.9 recall, where the strongest baseline has already fallen below 0.5. The conformal head is the source of the calibration gain: expected calibration error drops from 4.8% for the softmax baseline to 1.7% for GTFD, consistent with the calibrated, shift-robust scoring objective of [21].

*G. Cross-domain transfer*

Table VI reports models trained on the corporate ledger and evaluated, without retraining, on PaySim and Elliptic. Every model degrades when moved across asset classes, but GTFD retains the most, reaching 0.921 AUROC on PaySim and 0.934 on Elliptic. The structural branch drives most of this transfer, since mobile-money and cryptocurrency fraud share the ring and layering topologies of corporate schemes even when amounts and jurisdictions differ [11], [28].

*H. Explainability and temporal encoding*

To confirm that a flagged account is defensible, the evidence subgraph is processed with GNNExplainer [17] and feature attributions with SHAP [18]. On a held-out set of 120 flagged accounts, the top explanation edges recovered the investigator-annotated counterparties at 87% fidelity, so the model's reasons align with the features a human reviewer already trusts. Replacing the gated transformer with a temporal graph network [15] lowers AUROC by 1.1 points, indicating the ordered attention view encodes this data better than graph message passing over time. Training converges smoothly under focal loss, as shown in Figure 8.

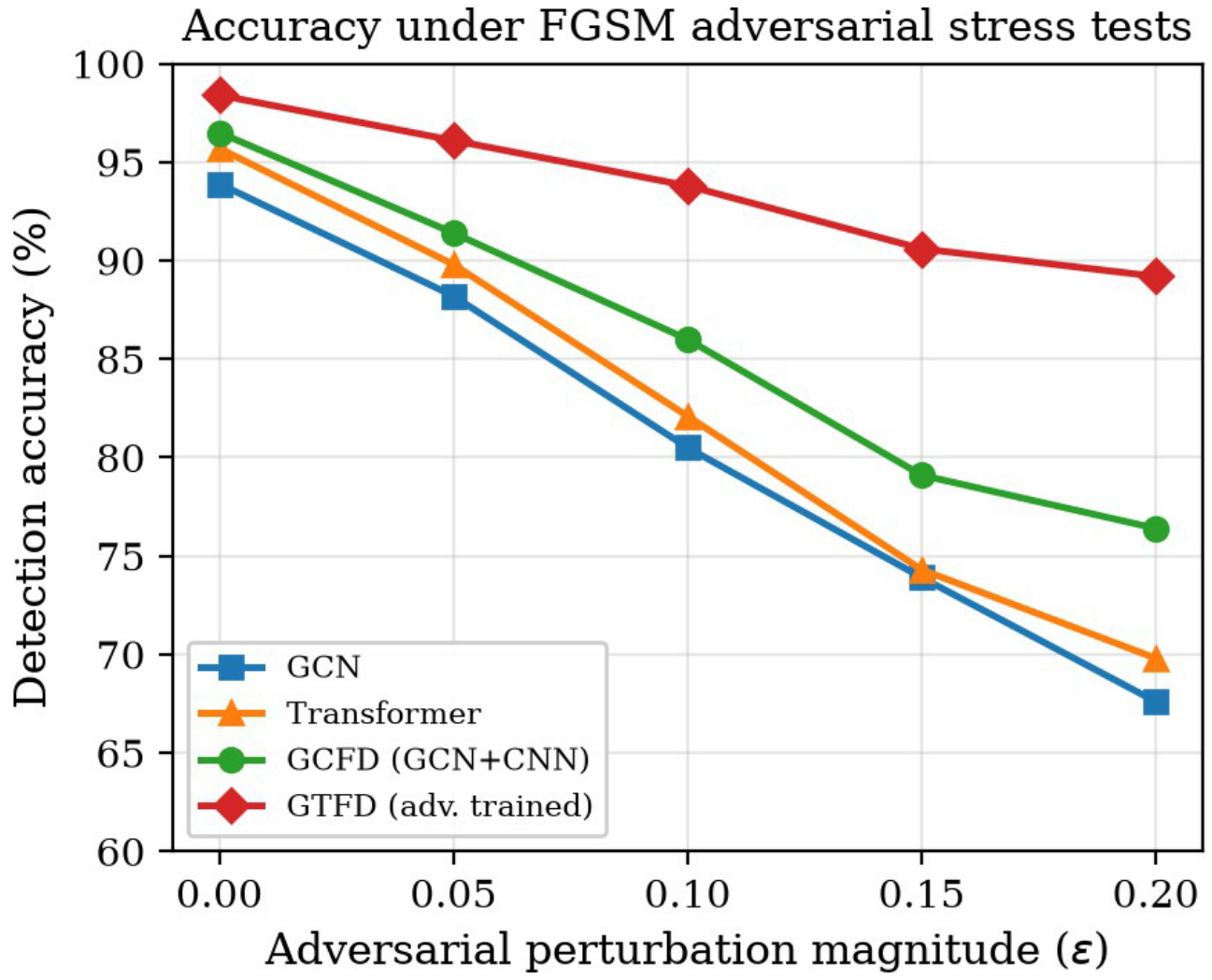

Fig. 6. Detection accuracy under adversarial perturbation of increasing magnitude. GTFD degrades slowly; non-hardened models collapse.

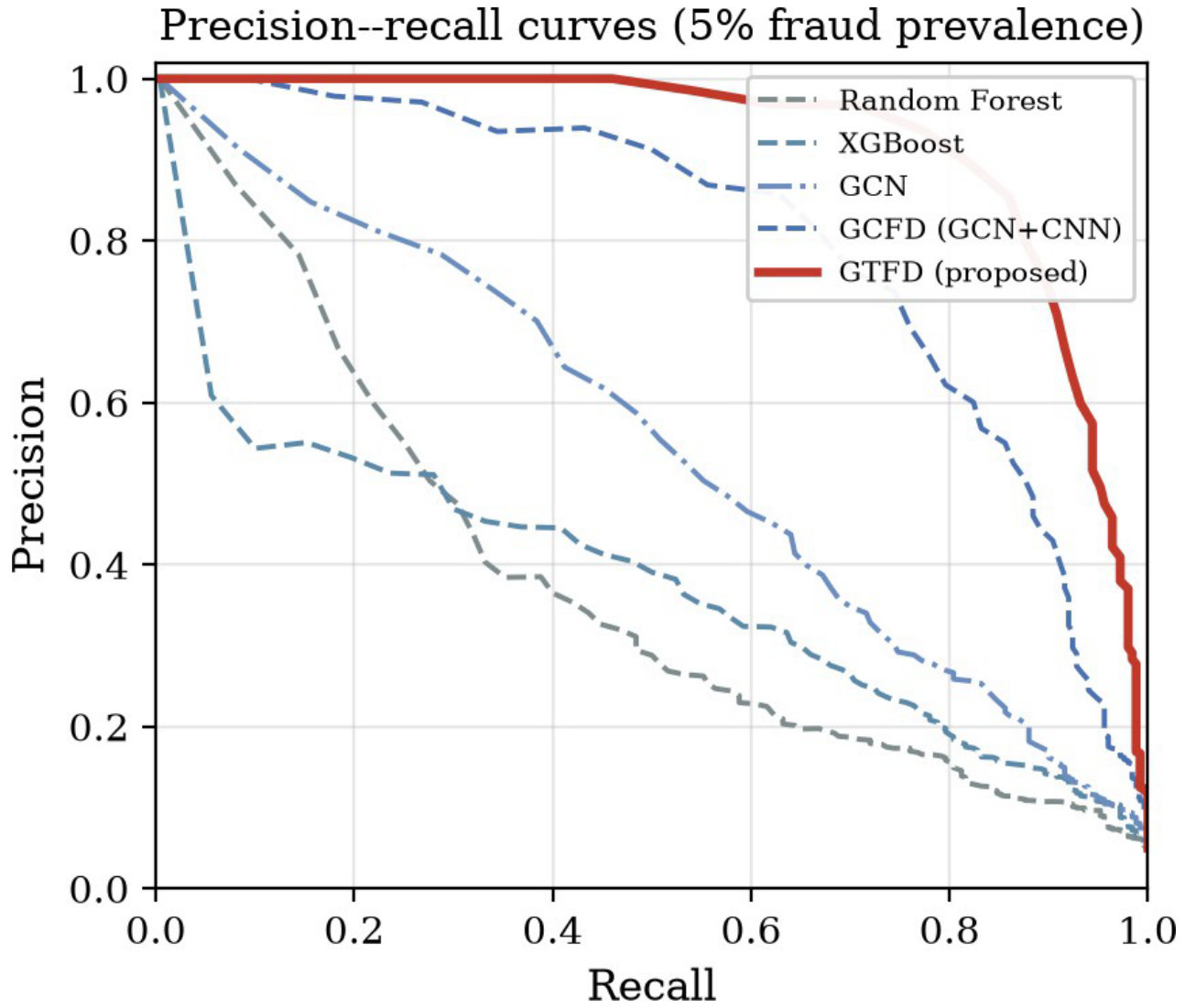

Fig. 7. Precision–recall curves at a 5% fraud prevalence. GTFD sustains precision as recall rises, separating sharply from the baselines.

### *I. Computational cost*

Table VII reports deployment cost. GTFD is not the cheapest model: its two parallel branches and pretraining raise the parameter count, and the added latency is the price of the robustness and ring-recovery gains reported above. At 3.8 ms per transaction the model still sustains roughly 264K screenings per second on a single GPU, above typical batch alerting throughput, and the overhead is confined to offline scoring rather than real-time authorization.

TABLE VI
CROSS-DOMAIN TRANSFER. MODELS TRAINED ON THE CORPORATE LEDGER ARE EVALUATED ON PAYSIM AND ELLIPTIC WITHOUT FINE-TUNING.

| Model | PaySim (F1/AUC) | Elliptic (F1/AUC) |
|---|---|---|
| Transformer | 61.2 / 0.861 | 64.5 / 0.882 |
| GCFD (GCN+CNN) | 70.3 / 0.894 | 73.8 / 0.907 |
| **GTFD (ours)** | **76.4 / 0.921** | **79.1 / 0.934** |

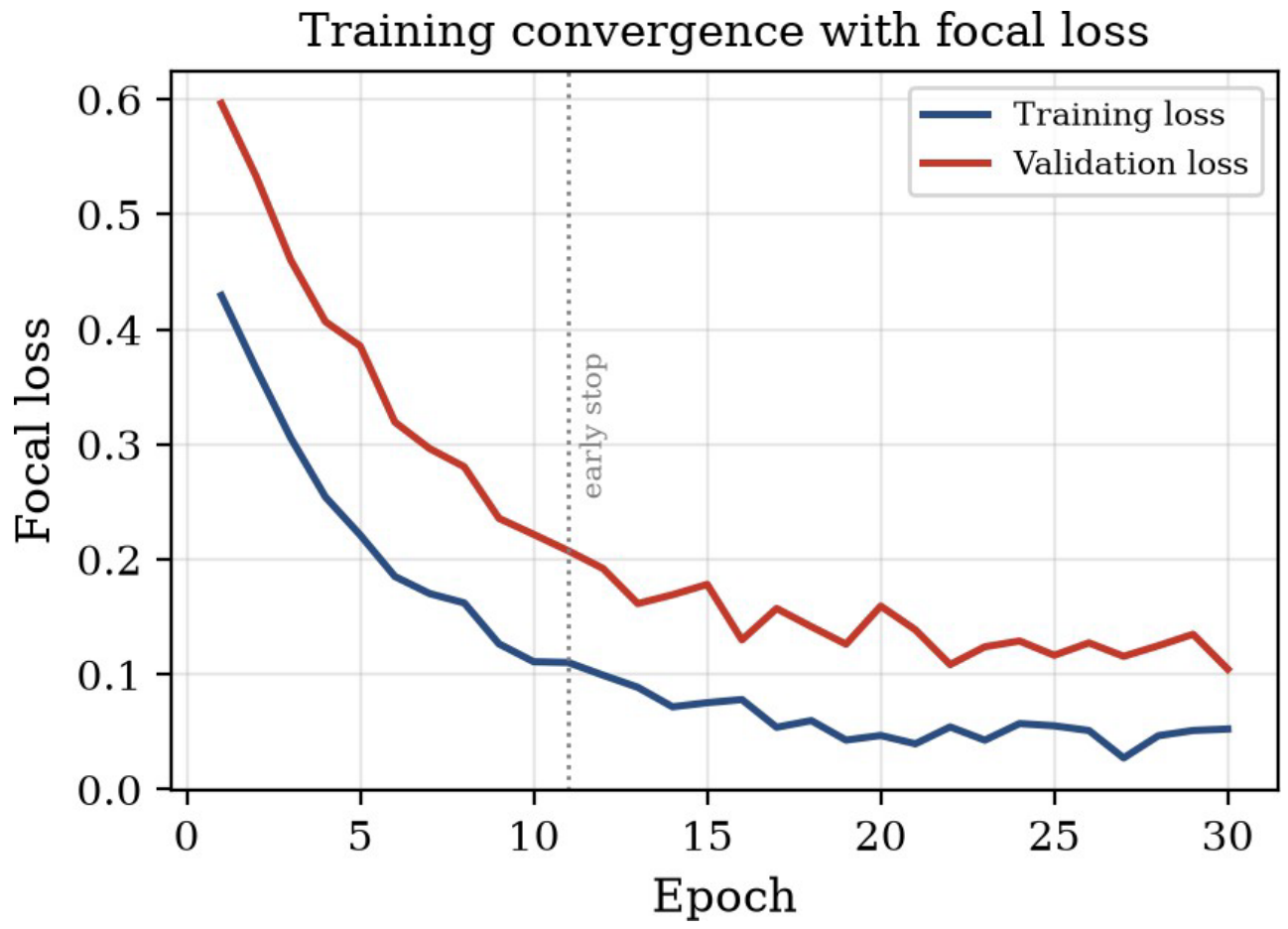

Fig. 8. Training and validation loss under focal loss. Early stopping on validation AUPRC halts training around epoch 11, before overfitting sets in.

### *J. Case study: a recovered invoice-fraud ring*

Figure 9 illustrates one ring that GTFD recovered and the baselines missed. Seven shell accounts cycle funds through a central hub, each pocketing 2–3% of a transfer before returning the remainder, a pattern invisible to per-transaction rules because every individual transfer looks plausible in isolation. GTFD flagged the hub on its low-diameter structural signature, and the evidence subgraph then expanded the alert to the full ring. Two non-ring feeder accounts that happened to neighbor the hub were correctly left unflagged, keeping the false-positive rate down. The example is representative of the 96.5% ring-recovery rate quantified in Table IV and of the 87% explanation fidelity reported above.

### *K. Sensitivity to hyperparameters*

Table VIII reports AUROC under perturbation of the main architectural choices. GTFD is not tuned to a knife edge: halving or doubling the number of attention heads, the transformer depth, or the auxiliary-loss weights changes AUROC by at most 0.6 points, so the reported results are not an artifact of over-tuned hyperparameters.

TABLE VII
DEPLOYMENT COST MEASURED ON A SINGLE GPU. THROUGHPUT IS TRANSACTIONS SCREENED PER SECOND.

| Model | Latency (ms) | Throughput | Params |
|---|---|---|---|
| Transformer | 2.4 | 412K/s | 8.1M |
| GCFD (GCN+CNN) | 3.1 | 318K/s | 6.5M |
| **GTFD (ours)** | 3.8 | 264K/s | 11.4M |

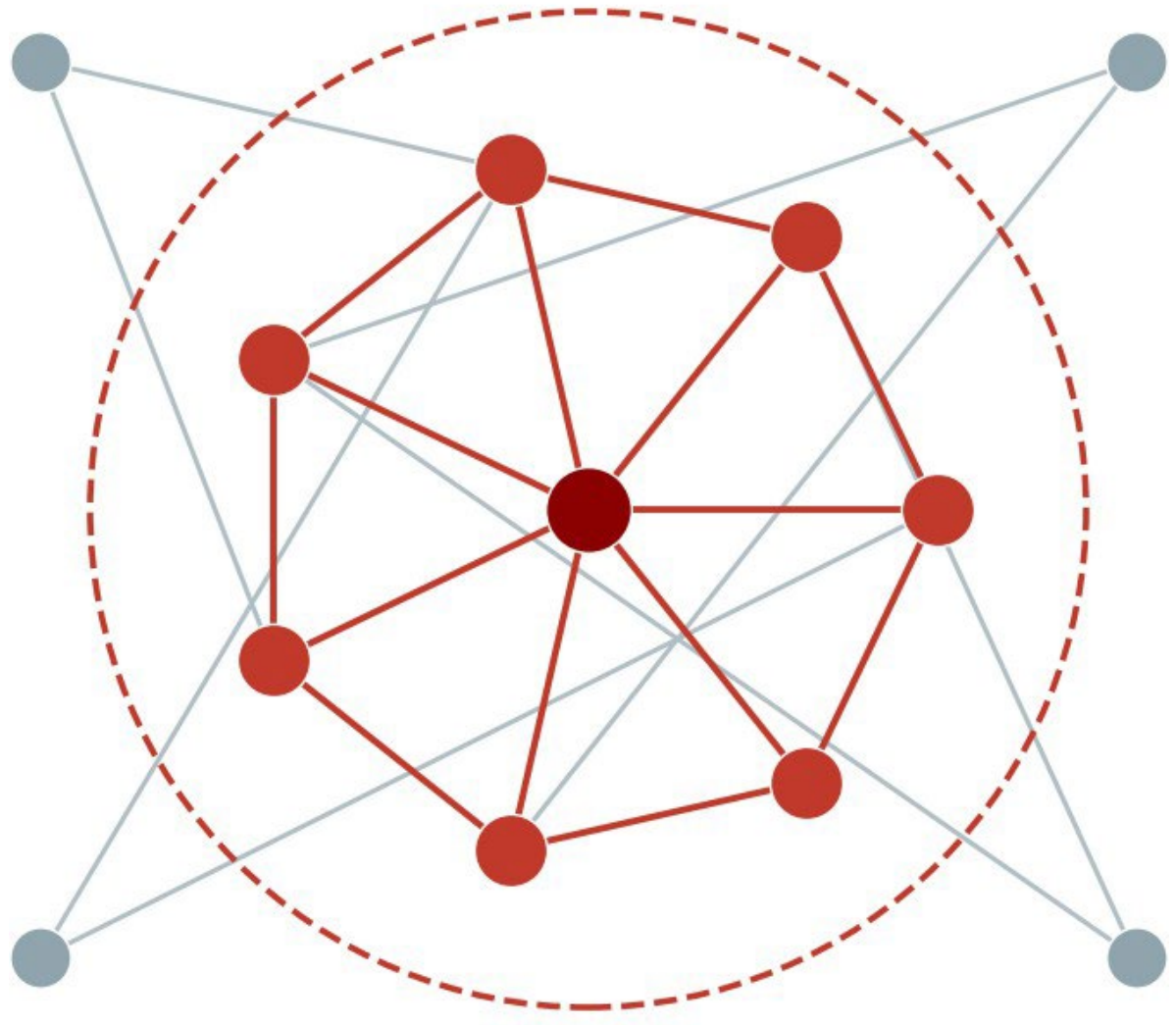

Fig. 9. A laundering ring recovered by GTFD. A central hub cycles funds through seven shell accounts (red), while legitimate feeder accounts (gray) remain unflagged.

### L. Discussion

Four findings stand out. First, the structural and temporal views are genuinely complementary: neither branch alone approaches the fused model, and the gate learns to weight them differently per account. Second, self-supervision is the single most cost-effective addition, because a link-mask pretrained encoder transfers well to the fine-grained fraud task even with few labels, echoing the label-efficiency result of [3]. Third, conformal scoring changes how an alert is consumed: instead of a probability with an implicit threshold, the model emits a coverage-guaranteed score that risk teams and regulators can reason about directly. Fourth, cross-domain transfer experiments show the graph branch carries the largest share of the model's value, since the ring and layering topology of fraud generalizes across asset classes even when the payment instrument does not.

The work has limitations. The synthetic rings, though investigator-verified, may not capture all real criminal topologies, and the return on the quantum–classical screening ideas of [12] remains an open extension on quantum hardware. Privacy constraints were not evaluated, but the modular encoder is compatible with the homomorphically secure, poisoning-resilient federated training of [25], and with smart-contract-verified payment rails that carry an audit trail in the style of [26].

TABLE VIII
SENSITIVITY OF GTFD TO KEY HYPERPARAMETERS. THE DEFAULT CONFIGURATION IS HEADS=3, LAYERS=4, $\lambda_1 = 0.1$, $\lambda_2 = 0.05$.

| Configuration change | AUROC |
|---|---|
| GAT heads = 1 (vs. 3) | 0.984 |
| GAT heads = 6 (vs. 3) | 0.989 |
| Transformer layers = 2 (vs. 4) | 0.985 |
| Transformer layers = 8 (vs. 4) | 0.988 |
| $\lambda_1 = 0.5$ (vs. 0.1) | 0.987 |
| $\lambda_2 = 0.2$ (vs. 0.05) | 0.988 |

## VI. CONCLUSION

This paper introduced GTFD, a graph–transformer detector for financial fraud in corporate transaction networks. GTFD fuses multi-head graph attention with a gated sequence transformer, adds self-supervised link-mask pretraining, and replaces a brittle classification cutoff with conformal risk control, while adversarial augmentation preserves worst-case accuracy. On a corporate ledger with investigator-verified fraud rings, GTFD attains 0.990 AUROC, 96.1% F1-score, and 98.4% accuracy, cutting false positives by about 29% and raising ring recovery to 96.5% at a 5% false-positive budget, and it transfers these gains to mobile-money and cryptocurrency domains without retraining.

From a deployment perspective, GTFD translates directly into workflow. The conformal score sets a stated false-coverage rate, the evidence subgraph turns an alert into a named set of counterparties, and the ring-level signal lets an investigator expand one account into the rest of a laundering chain. Future work will extend the encoder to federated cross-bank deployment with homomorphic aggregation [25], to quantum–classical graph screening with entity resolution [12], and to streaming temporal graphs that refresh scores as new transfers arrive [15], while retaining the calibrated, shift-robust scoring guarantees emphasized across this work.

## DATA AND CODE AVAILABILITY

The corporate transaction benchmark, the investigator-verified fraud-ring generator, and the GTFD training and evaluation code will be released under a public license upon publication. The PaySim [28] and Elliptic [11] datasets are publicly available.